\def\ANON{0} 

\ifnum\ANON=1
    \documentclass[sigconf,anonymous]{acmart}
\else
    \documentclass[sigconf]{acmart}
\fi

\usepackage{amssymb}
\usepackage{amsmath}
\usepackage{amsthm}
\usepackage{booktabs}
\usepackage{enumitem}
\usepackage{graphicx}
\usepackage{color}
\usepackage{multirow}
\usepackage{subcaption}
\usepackage{caption}
\AtBeginDocument{%
  \providecommand\BibTeX{{%
    \normalfont B\kern-0.5em{\scshape i\kern-0.25em b}\kern-0.8em\TeX}}}

\copyrightyear{2025}
\acmYear{2025}
\setcopyright{cc}
\setcctype{by}
\acmConference[ICAAI 2025]{2025 9th International Conference on Advances in Artificial Intelligence}{November 14--16, 2025}{Manchester, United Kingdom}
\acmBooktitle{2025 9th International Conference on Advances in Artificial Intelligence (ICAAI 2025), November 14--16, 2025, Manchester, United Kingdom}
\acmPrice{}
\acmDOI{10.1145/3787279.3787285}
\acmISBN{979-8-4007-2104-5/2025/11}

\begin{document}

\title[Bayesian Optimization for Portfolio Management with Adaptive Scheduling]{Improving Bayesian Optimization for Portfolio Management with an Adaptive Scheduling}

\author{Zinuo You}
\orcid{0009-0003-7840-5999}
\affiliation{%
    \department{School of Computer Science}
    \institution{University of Bristol}
    \country{Bristol, UK}
}
\email{zinuo.you@bristol.ac.uk}
\author{John Cartlidge}
\orcid{0000-0002-3143-6355}
\thanks{Corresponding Author. Email: john.cartlidge@bristol.ac.uk.}
\affiliation{%
    \department{School of Engineering Mathematics and Technology}
    \institution{University of Bristol}
    \country{Bristol, UK}
}
\email{john.cartlidge@bristol.ac.uk}
\author{Karen Elliott}
\orcid{0000-0002-2455-0475}
\affiliation{%
    \department{Business School}
    \institution{University of Birmingham}
    \country{Birmingham, UK}
}
\email{k.elliott@bham.ac.uk}
\author{Menghan Ge}
\orcid{0009-0005-6950-1756}
 \affiliation{%
    \institution{Stratiphy Limited}
    \country{London, UK}
}  
\email{menghan.ge@stratiphy.io}
\author{Daniel Gold}
\orcid{0009-0002-9747-7595}
 \affiliation{%
    \institution{Stratiphy Limited}
    \country{London, UK}
}  
\email{daniel@stratiphy.io}

\renewcommand{\shortauthors}{You and Cartlidge, et al.}






\begin{abstract}
Existing black-box portfolio management systems are prevalent in the financial industry due to commercial and safety constraints, although their performance can fluctuate dramatically with changing market regimes. Evaluating these non-transparent systems is computationally expensive, as fixed budgets limit the number of possible observations. Therefore, achieving stable and sample-efficient optimization for these systems has become a critical challenge. This work presents a novel Bayesian optimization framework (TPE-AS) that improves search stability and efficiency for black-box portfolio models under these limited observation budgets. Standard Bayesian optimization, which solely maximizes expected return, can yield erratic search trajectories and misalign the surrogate model with the true objective, thereby wasting the limited evaluation budget. To mitigate these issues, we propose a weighted Lagrangian estimator that leverages an adaptive schedule and importance sampling. This estimator dynamically balances exploration and exploitation by incorporating both maximization of model performance and minimization of the variance of model observations. It guides the search from broad, performance-seeking exploration towards stable and desirable regions as the optimization progresses. Extensive experiments and ablation studies, which establish our proposed method as the primary approach and other configurations as baselines, demonstrate its effectiveness across four backtest settings with three distinct black-box portfolio management models.
\end{abstract}
\maketitle

\section{Introduction}
Stock portfolio management has been an open issue in academia and industry due to its inherent complexity and evolving nature. The challenge is to select an optimal set of stocks and determine the optimal weights for each stock to maximize portfolio returns. Traditional portfolio management methods, such as the Markowitz approach, prioritize asset diversification to mitigate risk by balancing asset correlations~\cite{zhang2018portfolio}. However, these methods have restrictive prior assumptions, such as normally distributed returns and stationary markets, which contradict the evolving nature of market conditions~\cite{kulali2016portfolio,zhang2018portfolio}. Consequently, research efforts have shifted toward the capture of complex non-linear dynamics in stock markets using data-driven methods~\cite{krishnamoorthy2023diffusion,li2021openbox}. These efforts typically focus on designing novel trading strategies. However, in real-world industrial applications, portfolio management often relies on non-transparent trading models (i.e., proprietary black-box systems). These models are generally expensive to evaluate because of computational costs and safety issues. Consequently, Bayesian optimization has become a powerful framework for optimizing these black-box systems. Unlike traditional stochastic methods (e.g., particle swarm optimization \cite{zhu2011particle}), Bayesian optimization can efficiently navigate the exploration-exploitation trade-off in non-convex and noisy environments, making it ideal for high-stakes financial applications. It consists of a surrogate model and an acquisition function, which approximates the unknown objective and finds the optimum parameters. However, when applied to complex and noisy black-box functions with a severely limited evaluation budget, conventional Bayesian optimization can suffer from an inefficient search process. By focusing solely on maximizing the expected performance, optimization can become unstable, producing erratic trajectories, which also misaligns the surrogate's target distribution with the true distribution. This instability reveals a fundamental flaw: an objective function focused solely on maximizing performance is ill-suited for noisy, budget-constrained financial model tuning. Therefore, we argue that the objective itself must be redefined to include stability of performance as a critical consideration.

To address these challenges, we propose a novel Bayesian optimization framework featuring an adaptive weighted Lagrangian estimator. Our approach first reframes the optimization as a bi-objective problem that explicitly balances maximizing model performance with minimizing the variance of the observed performance scores. Then, we adopt an adaptive Lagrangian schedule to shift the optimization focus over time. Finally, we leverage clipped importance sampling to correct the surrogate bias with respect to the underlying true distribution. The contribution of this work is threefold:
\begin{itemize}
    \item We propose a novel Bayesian optimization framework that improves search stability and model performance through a bi-objective estimator with an adaptive scheduling, explicitly balancing performance maximization and observation variance minimization.
    \item We demonstrate that the proposed TPE-AS consistently outperforms baselines over three distinct black-box portfolio models across four real-world backtest settings.
    \item We show that the proposed adaptive scheduling Bayesian optimization leads to a more reliable search of high-quality parameter configurations than the conventional setting.
\end{itemize}

\noindent
Code is available online: \href{https://github.com/pixelhero98/TPE-AS}{https://github.com/pixelhero98/TPE-AS}.

\section{Related Work}
\subsection{Portfolio Optimization Models}
The method of managing stock portfolios has evolved significantly from Markowitz's foundational mean-variance theory \cite{markowitz1952modern}, which laid the groundwork with assumptions of normality and stationarity. To address these limitations, Stochastic Portfolio Optimization (SPT) emerged, explicitly modeling uncertainty in asset returns. This paradigm incorporates sophisticated risk measures such as Conditional Value-at-Risk (CVaR) to capture tail risk~\cite{kaut2007stability,sarykalin2008value}. Metaheuristics (e.g., GAs \cite{metaxiotis2012multiobjective}) and, more recently, advanced data-driven models from reinforcement learning and deep learning \cite{alzaman2025optimizing,wang2021deeptrader,cheng2024optimizing} have focused on new trading strategies rather than efficient tuning of existing black-box systems. However, the critical industrial problem of efficiently tuning the hyperparameters of existing, non-transparent (i.e., black-box) systems remains a significant challenge.

\subsection{Bayesian Optimization}
Bayesian optimization is a well-established framework for sample-efficient optimization of expensive-to-evaluate black-box functions, particularly when gradient information is unavailable \cite{shahriari2015taking}. Common surrogate models include Gaussian Processes (GPs) \cite{jones1998efficient,bull2011convergence}, Tree-structured Parzen Estimators (TPEs) \cite{ozaki2022multiobjective}, and Bayesian Neural Networks (BNNs) \cite{kerleguer2024bayesian}, which are paired with acquisition functions like Expected Improvement (EI) or Upper Confidence Bound (UCB) to navigate the fundamental trade-off between exploration and exploitation \cite{snoek2012practical,kaufmann2012bayesian}. However, under severe noise and scarce evaluation samples, which commonly exist in financial market applications, standard Bayesian optimization often bounces among poor regions and fails to converge to desired outcomes. A singular focus on maximizing expected performance can lead to an erratic search trajectory, which wastes evaluations and fails to converge to desired solutions.

\section{Problem Formulation}
Consider a portfolio management model $f$, which accepts a hyperparameter configuration $\mathbf{x}=\{x_1,x_2,\ldots,x_m\}$ from a bounded domain $\mathcal{X}\in \mathbb{R}^m$. The function $f(\mathbf{x})$ returns a scalar value (e.g., annualized Sharpe ratio) representing the performance of the portfolio strategy. Here, $\mathbf{x}=\{x_1,x_2,\ldots,x_m\}$ consists of a set of mixed integer and continuous hyperparameters that define the behavior of a given trading strategy (in all settings, the number of hyperparameters $m\geq25$). The conventional objective in Bayesian optimization is to find the global maximizer of the black-box function, given by:
\begin{equation}
    \mathbf{x}^*=\arg \underset{\mathbf{x} \in \mathcal{X}}{\max} \;\mathbb{E}[f(\mathbf{x})].
    \label{conobj}
\end{equation}

As stated previously, pursuing this single objective directly can lead to an unstable and inefficient search process, as we do not have access to the internal function of the trading model and must operate within a strict evaluation budget $\eta$. Therefore, to promote a more stable and efficient search, we formulate the hyperparameter tuning of the portfolio management model as a stochastic bilevel optimization problem. Rather than optimizing $\mathbb{E}(f(\mathbf{x}))$, we treat $f(\mathbf{x})$ as a noisy Sharpe ratio evaluator under a strict budget $\eta$. Therefore, we seek:
\begin{align}
\max_{\mathbf{x} \in \mathcal{X}} \mathbb{E}(f(\mathbf{x})),\;\min_{\mathbf{x} \in \mathcal{X}} \sigma^2(f(\mathbf{x})),
\label{dual-obj}
\end{align}
which captures both model performance and performance stability.

\section{Methodology}\label{sec:method}
The proposed method, TPE-AS, focuses on improving the stability and expected performance of hyperparameter tuning for expensive black-box portfolio models with limited evaluations. As $m\geq25$ and often involves both discrete and categorical hyperparameters, we select TPE as the surrogate model and EI as the acquisition function, with an adaptive weighted Lagrangian estimator as the overall objective function $\mathcal{J}$.

\subsection{The TPE framework}
The TPE is a sequential model-based optimization method that works by modeling two separate probability densities over the hyperparameter space. TPE partitions the history of evaluated points $\mathcal{D}_n$ based on an objective score threshold $\mathcal{J}^*$, which is the $k$-quantile (e.g., $k=15\%$) of the observed past performance scores. This creates two groups:
\begin{itemize}
    \item A good group, $\gamma(\mathbf{x})$, modeling the distribution of hyperparameters for high-performing observations, such that $\gamma(\mathbf{x}) = P\left(\mathbf{x} |\mathcal{J}(\mathbf{x}) \geq \mathcal{J}^*\right)$.
    \item A bad group, $\Omega(\mathbf{x})$, modeling the distribution of hyperparameters for lower-performing observations, such that $\Omega(\mathbf{x}) = P\left(\mathbf{x} |\mathcal{J}(\mathbf{x}) < \mathcal{J}^*\right)$.
\end{itemize}
TPE fits these densities using kernel density estimators, which avoids making strong functional assumptions and handles mixed-type (continuous and discrete) hyperparameters efficiently. The acquisition function is then defined by the ratio of these two densities. The next point to evaluate, $\mathbf{x}_{n+1}$, is chosen to maximize,
\begin{equation}
    \alpha(\mathbf{x}) = \frac{\gamma(\mathbf{x})}{\Omega(\mathbf{x})}.
\end{equation}
By maximizing this ratio, the search is guided toward regions where hyperparameters are likely to produce high performance (high $\alpha(\mathbf{x})$) and unlikely to produce low performance (low $\alpha(\mathbf{x})$). This non-parametric approach effectively balances exploration and exploitation and is well-suited for high-dimensional (high $m$) financial tuning problems.

\begin{table*}[t]
\centering
\caption{Comparison of maximum model performance ($\mathbb{E}(f(x))$ annualized Sharpe ratio) for the proposed method and baselines across twelve scenarios ($\eta=500$). A larger Sharpe ratio indicates a more profitable portfolio.}
\begin{tabular*}{\textwidth}{@{\extracolsep{\fill}} l|cccc|cccc|cccc @{}}
\toprule
\toprule
\multirow{2}{*}{\textbf{Method} }& \multicolumn{4}{c|}{\textbf{M1}} & \multicolumn{4}{c|}{\textbf{M2}} & \multicolumn{4}{c}{\textbf{M3}} \\
 & \textbf{S1} & \textbf{S2} & \textbf{S3} & \textbf{S4} & \textbf{S1} & \textbf{S2} & \textbf{S3} & \textbf{S4} & \textbf{S1} & \textbf{S2} & \textbf{S3} & \textbf{S4} \\
\hline
GP-EI   & 2.998 & 1.904 & 1.028 & 0.719 & 2.976 & 1.672 & \textbf{1.222} & \textbf{0.860} & 3.288 & 1.596 & 1.040 & 0.783 \\
GP-UCB  & 2.752 & 1.656 & 0.965 & 0.545 & 2.730 & 1.424 & 1.159 & 0.686 & 3.192 & 1.386 & 0.977 & 0.594 \\
GP-PI   & 2.866 & 1.665 & 1.001 & 0.699 & 2.844 & 1.433 & 1.195 & 0.840 & 3.323 & 1.394 & 1.013 & 0.761 \\
BNN-EI  & 2.999 & \textbf{2.001} & 0.989 & 0.689 & 3.145 & 1.849 & 1.093 & 0.776 & 3.479 & 1.660 & 1.001 & 0.751 \\
BNN-UCB & 2.646 & 1.955 & 0.998 & 0.568 & 2.792 & 1.920 & 1.102 & 0.655 & 3.070 & 1.565 & 1.010 & 0.619 \\
BNN-PI  & 2.665 & 1.547 & 0.981 & 0.711 & 2.811 & 1.595 & 1.085 & 0.798 & 3.095 & 1.295 & 0.993 & 0.775 \\
\hline
\textbf{TPE-AS} & \textbf{3.091} & 1.678 & \textbf{1.035} & \textbf{0.754} & \textbf{3.405} & \textbf{2.006} & {1.049} & {0.787} & \textbf{3.480} & \textbf{1.673} & \textbf{1.047} & \textbf{0.821} \\
\bottomrule
\bottomrule
\end{tabular*}
\label{perf-eval}
\end{table*}

\begin{table*}[t]
\centering
\caption{Comparison of model performance variance ($\sigma^{2}(f(x))$ optimization stability) for the proposed method and baselines across twelve scenarios ($\eta=500$). Lower variance indicates a more stable optimization process.}
\begin{tabular*}{\textwidth}{@{\extracolsep{\fill}} l|cccc|cccc|cccc @{}}
\toprule
\toprule
\multirow{2}{*}{\textbf{Method} }& \multicolumn{4}{c|}{\textbf{M1}} & \multicolumn{4}{c|}{\textbf{M2}} & \multicolumn{4}{c}{\textbf{M3}} \\
 & \textbf{S1} & \textbf{S2} & \textbf{S3} & \textbf{S4} & \textbf{S1} & \textbf{S2} & \textbf{S3} & \textbf{S4} & \textbf{S1} & \textbf{S2} & \textbf{S3} & \textbf{S4} \\
\hline
GP-EI   & 1.278 & 0.407 & 0.180 & 0.078 & 0.900 & 0.277 & 0.171 & 0.085 & 1.327 & 0.220 & 0.176 & 0.052 \\
GP-UCB  & 1.382 & 0.497 & 0.191 & 0.102 & 1.004 & 0.367 & 0.182 & 0.109 & 1.435 & 0.268 & 0.187 & 0.068 \\
GP-PI   & 0.690 & 0.223 & 0.116 & 0.110 & 0.512 & 0.193 & 0.107 & 0.066 & 0.717 & 0.120 & 0.113 & 0.039 \\
BNN-EI  & 1.917 & 1.106 & 0.453 & 0.217 & 1.667 & 1.070 & 0.468 & 0.221 & 1.991 & 0.597 & 0.443 & 0.145 \\
BNN-UCB & 1.955 & 1.119 & 0.697 & 0.222 & 1.705 & 1.083 & 0.712 & 0.226 & 2.031 & 0.604 & 0.148 & 0.148 \\
BNN-PI  & 1.341 & 0.533 & 0.209 & 0.096 & 1.091 & 0.497 & 0.224 & 0.114 & 1.393 & 0.288 & 0.204 & 0.073 \\
\hline
\textbf{TPE-AS} & \textbf{0.613} & \textbf{0.131} & \textbf{0.073} & \textbf{0.049} & \textbf{0.492} & \textbf{0.170} & \textbf{0.111} & \textbf{0.049} & \textbf{0.309} & \textbf{0.084} & \textbf{0.071} & \textbf{0.033} \\
\bottomrule
\bottomrule
\end{tabular*}
\label{var-eval}
\end{table*}

\subsection{Adaptive Weighted Lagrangian Estimator}
\subsubsection{Dual Objectives}\label{sec:method:bi-objective} 
The portfolio optimization problem in Eq.~(\ref{dual-obj}) presents a dual challenge. Direct optimization of these two objectives typically requires multistage procedures, which are undesirable and impractical in evolving financial decision-making due to their computational complexity and lack of unified optimization criteria. Moreover, Bayesian optimization frameworks are most efficient when applied to a single objective. 

To reconcile these issues, we present an adaptive Lagrangian estimator that balances these two objectives. Accordingly, we reformulate Eq.~(\ref{dual-obj}) into a constrained optimization problem using a Lagrangian multiplier (e.g., see \cite{ito2008lagrange}),
\begin{equation} \max\;\mathcal{J}(\mathbf{x},\lambda) = \max\;\{\mathbb{E}[f(\mathbf{x})]- \lambda\sigma^2(f(\mathbf{x}))\}, \label{lag}
 \end{equation} 
where $\lambda$ is the regularization term. This formulation explicitly enforces stable optimization trajectories while maximizing expected model performance. Furthermore, to promote a more stable search process, we adopt an adaptive schedule for the penalty term $\lambda$ at each step $t$. In this sense, we start with $\lambda_t \approx 0$ to prioritize expected portfolio returns and then ramp $\lambda_t \rightarrow 1$ to penalize model performance variance as the budget depletes. The scheduling is defined by
\begin{equation}
    \lambda_t=\frac{1-\cos{(\min(\frac{t}{\eta}\pi,\pi))}}{2},\; t = 1,2,...,\eta.
\end{equation}

\subsubsection{Importance Sampling}\label{sec:method:sampling} 
To address the potential divergence between the predictive distribution of the surrogate model (the proposal distribution $q_{\theta}(\mathbf{x})$, that is $\gamma(\mathbf{x})$), and the true underlying distribution of high-quality portfolios (the true distribution $g(\mathbf{x})$), we incorporate importance sampling. This allows us to correct for distributional shifts during optimization. The full objective function, which incorporates the importance weight, becomes
\begin{equation} 
\mathcal{J}(x,\lambda) =\mathbb{E}_{q_{\theta}}[f(x)]-\lambda_t\sigma_{q_{\theta}}^{2}(f(x)\frac{g(x)}{q_{\theta}(x)}), 
\label{interobj}
\end{equation}
where the expectation and the variance term are taken with respect to the proposal distribution $q_{\theta}(\mathbf{x})$. In this sense, Eq.~(\ref{interobj}) enables a principled trade-off between achieving high performance and controlling risk under distributional shifts.

In practice, the importance weight can vary considerably. To mitigate this, we apply a clipping mechanism, resulting in the final form of our estimator,
\begin{equation}
 \mathcal{J}(x,\lambda) = \mathbb{E}_{q_{\theta}}[f(x)]-\lambda_t\sigma_{q_{\theta}}^{2}(f(x)\cdot \text{clip}(\frac{g(x)}{q_{\theta}(x)}, 1-\epsilon, 1+\epsilon)) .
    \label{finalobj}
\end{equation}
Here, $\epsilon$ controls the bias-variance trade-off of the importance weight correction, normally $\epsilon \in [0.1,0.35]$. This objective guides the TPE surrogate to find solutions that are both high-performing and stable, and align with the desired target distribution.  

\begin{figure*}[tb]
    \centering
    \begin{subfigure}{0.8\linewidth}
        \includegraphics[width=\linewidth]{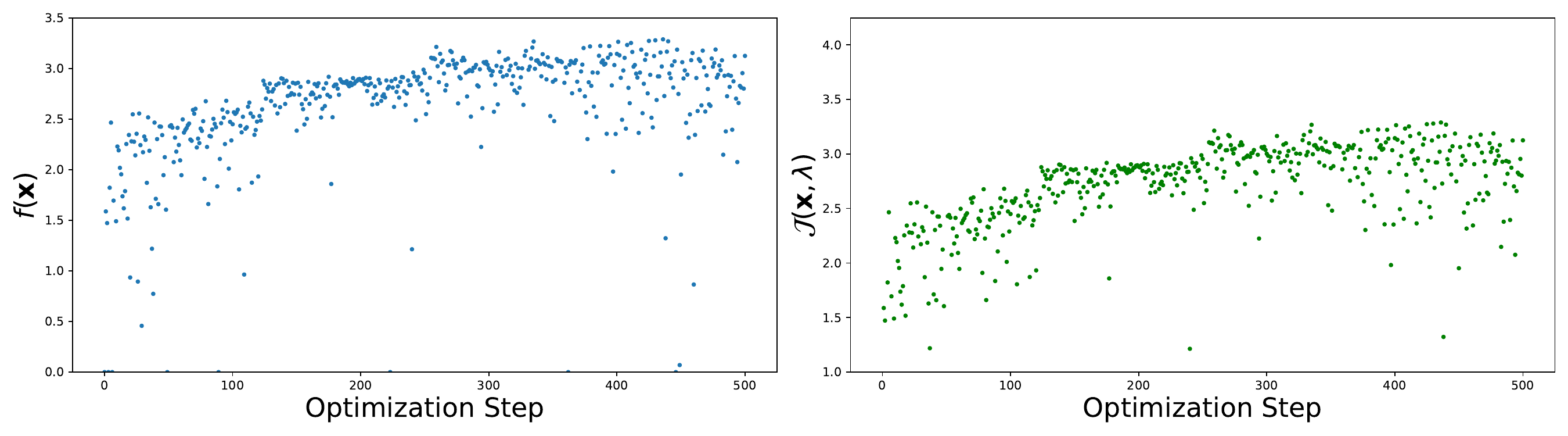}
        \caption{Trajectories of TPE guided by $\mathcal{J}(\mathbf{x},\lambda)$, where $\mathbb{E}(f(\mathbf{x}))=3.288$ (annualized Sharpe ratio = 3.288) and $\sigma^2_{q_{\theta}}(f(\mathbf{x}))=0.309$ (optimization stability).}
        \label{ab1}
    \end{subfigure}
    \begin{subfigure}{0.8\linewidth}
        \includegraphics[width=\linewidth]{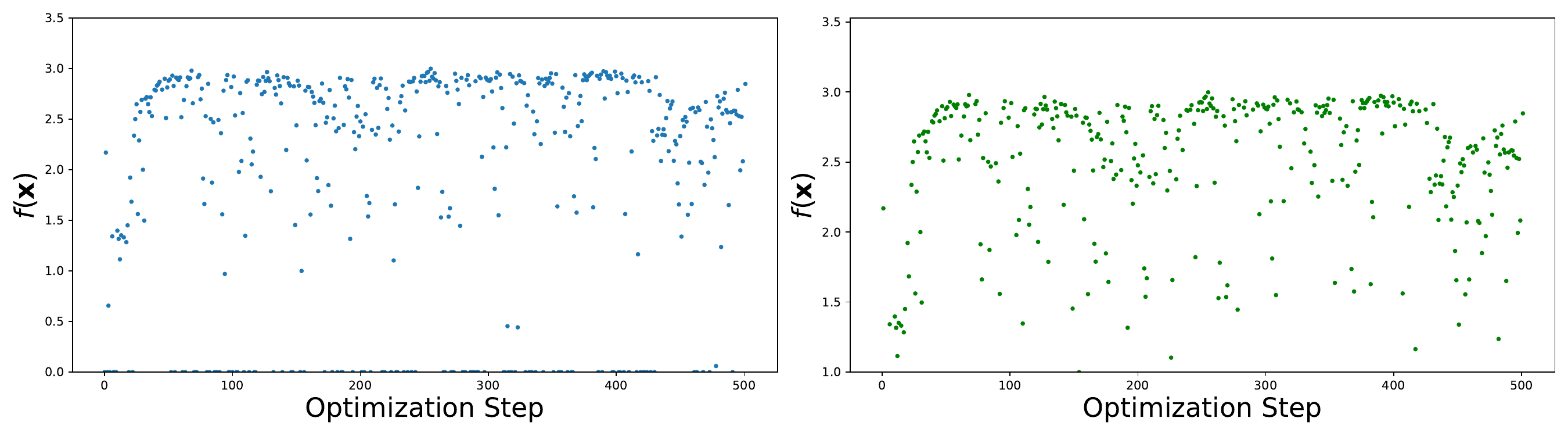} 
        \caption{Trajectories of TPE guided by $f(\mathbf{x})$, where $\mathbb{E}(f(\mathbf{x}))=2.980$ (annualized Sharpe ratio = 2.980) and $\sigma^2_{q_{\theta}}(f(\mathbf{x}))=1.367$ (optimization stability).}
        \label{ab2}
    \end{subfigure}
    \caption{The x-axis shows 500 optimization steps. Blue dots represent portfolio model performance; green dots represent the corresponding objective score for each optimization step.}
    \label{ablation}
\end{figure*}

\section{Experiments}\label{sec:results}
\subsection{Experimental Setting}
We conducted experiments on twelve scenarios with six baselines. These scenarios comprised three distinct black-box portfolio management models (M1-M3) tested in four different backtest settings (S1-S4). The observation budget was fixed at $\eta=500$. Due to commercial sensitivity imposed by the company, the details of the black-box models, the backtest settings, and hyperparameters are not disclosed. However, we provide conceptual descriptions below to ensure that the experiments are clear.

\ifnum\ANON=1  
    \newpage
\else
    \vspace{1em}
\fi
\noindent
\textbf{Baselines}
\begin{itemize}
    \item GP-EI, GP-UCB, GP-PI: These baselines use a GP as the surrogate model. GPs are a standard choice in Bayesian optimization. It becomes computationally expensive when the size of the hyperparameter space, $m$, increases.
    \item BNN-EI, BNN-UCB, BNN-PI: These baselines adopt a BNN as the surrogate. BNNs are powerful for capturing complex, non-stationary patterns by leveraging neural networks' learning ability.
\end{itemize}
\textbf{Black-box Model}
\begin{itemize}
    \item M1 (Trend-Following): A strategy that generates signals based on momentum indicators.
    \item M2 (Mean-Reversion): A strategy that identifies assets that have deviated significantly from their historical averages.
    \item M3 (Threshold-Based Hybrid): A complex, threshold-based hybrid strategy that generates trading signals only when specific conditions are met.
\end{itemize}
\textbf{Backtest Settings}
\begin{itemize}
    \item S1 (High-Volatility Market): Simulates a short period (approximately 1 year) with frequently changing patterns.
    \item S2 (Stable Bull Market): Simulates a period of (approximately 3 years) low volatility and consistent upward trends.
    \item S3 (Range-Bound Market): Simulates a `sideways' market with no clear long-term trend (approximately 5 years).
    \item S4 (Range-Bound Market): Simulates a `sideways' market with no clear long-term trend (approximately 4 years).
\end{itemize}
\textbf{Hyperparameter Configuration}
\begin{itemize}
    \item $\mathbf{x}=\{x_1,x_2,\ldots,x_m\}$ includes lookback periods (moving averages), buy/sell signal thresholds (relative strength index level), and risk and sizing parameters (stop-loss level).
\end{itemize}

\subsection{Performance Evaluation}
As shown in Table~\ref{perf-eval}, TPE-AS achieves the highest annualized Sharpe ratio in eight out of twelve test cases, outperforming both GP-based and BNN-based baselines. In the remaining scenarios, notably M2-S3 and M3-S4, its performance is only marginally below the top entry. Table~\ref{var-eval} further highlights optimization stability via the variance of achieved Sharpe ratios. Here, TPE-AS consistently exhibits the lowest variance across all twelve scenarios, with its maximum observed variance just 0.170 (in M2-S2) versus BNN-UCB's 1.083 under the same scenario. The contrast is especially stark in M3-S1 (Threshold-Based Hybrid, High-Volatility Market), where TPE-AS's variance is 0.309 compared to 2.031 for BNN-UCB. This is likely due to the BNN's tendency to overfit when trained on limited samples, leading to an unstable approximation of the objective function and an erratic search path. 

In addition, Table~\ref{time-eval} reports the average time per optimization iteration (including portfolio response). TPE-AS completes each step in 43.22–58.76 s, outpacing the nearest competitor, GP-UCB, by up to 1.77 s per step and offering a 10\%–30\% speed-up over BNN-based methods. Taken together, these results show that TPE-AS not only achieves superior or near-optimal peak performance but does so with markedly greater stability and lower runtime.

\begin{table}[t]
    \caption{Average time (in seconds) taken for each optimization step, including portfolio model response time.}
    \label{time-eval}
    \begin{tabular}{@{}l|cccc@{}}
        \toprule
        \toprule
        Method & M3-S1 & M3-S2 & M3-S3 & M3-S4 \\
        \midrule
        GP-EI    & 45.12 & 46.49 & 48.71 & 60.62\\
        GP-UCB   & 44.99 & 46.33 & 48.93 & 60.41 \\
        GP-PI    & 46.48 & 45.66 & 48.69 & 60.72 \\
        BNN-EI   & 64.19 & 69.55 & 68.5 & 77.81 \\
        BNN-UCB  & 63.57 & 70.90 & 69.17 & 77.16\\
        BNN-PI   & 64.25 & 71.99 & 69.62 & 77.96\\
        \hline
        \textbf{TPE-AS}  & 43.22 & 45.13 & 47.77 & 58.76 \\
        \bottomrule
        \bottomrule
    \end{tabular}
\end{table}

\subsection{Ablation Study}
We compare our TPE-AS (guided by \(\mathcal{J}(\mathbf{x},\lambda)\)) against a standard TPE that maximizes only \(f(\mathbf{x})\), under the most challenging M3-S1 setup (Threshold-Based Hybrid in a High-Volatility market). Figure~\ref{ab1} shows the trajectories of TPE-AS in 500 optimization steps: blue dots (portfolio Sharpe ratios) rapidly climb into a tight cluster, matching the low variance in Table~\ref{var-eval} (\(\sigma^2=0.309\)). While green dots (objective score \(\mathcal{J}\)) gradually impose a stronger variance penalty, steering the search toward a stable optimum. By contrast, Figure~\ref{ab2} illustrates the conventional TPE's trajectory: erratic blue spikes ($f(\mathbf{x})=0$) and deep performance drops frequently occur, which yield a much higher variance (\(\sigma^2=1.367\)). Although the baseline eventually reaches \(f(\mathbf{x})=2.980\), it falls short of TPE-AS' peak of 3.288, highlighting the benefit of variance-aware scheduling.

\section{Conclusions}\label{sec:conclusions}
In this work, we present TPE-AS, a novel Bayesian optimization framework that delivers sample-efficient hyperparameter tuning for costly black-box portfolio models. By augmenting the standard TPE algorithm with an adaptive Lagrangian estimator, TPE-AS explicitly trades off between chasing higher Sharpe ratios and maintaining search stability. Our experiments yield three key insights. First, TPE-AS produces markedly smoother and more reliable optimization trajectories than GP-based baselines or BNN-based baselines. Second, our ablation confirms that adaptive scheduling is essential: it curbs the erratic evaluations of a pure performance objective and keeps the search focused on robust regions. Third, among surrogate choices, TPE scales better than GPs and avoids the overfitting pitfalls of BNNs under tight evaluation budgets. 

\balance


\begin{acks}
\ifnum\ANON=1 
        Redacted for anonymous review.
\else
        This work was supported by UK Research and Innovation (UKRI) Innovate UK grant number 10094067: Stratlib.AI -- A trusted machine learning platform for asset and credit managers.    
\fi
\end{acks}

\bibliographystyle{ACM-Reference-Format}
\bibliography{shortrefs}


\end{document}